\documentclass[10pt,twocolumn,letterpaper]{article}

 \usepackage{cvpr}              
\usepackage{url}            
\usepackage{booktabs}       
\usepackage{amsfonts}       
\usepackage{nicefrac}       
\usepackage{microtype}      
\usepackage{xcolor}         

\usepackage{graphicx}
\usepackage{booktabs}
\usepackage[accsupp]{axessibility}
\usepackage{multirow}
\usepackage{amssymb}
\usepackage[linesnumbered,ruled,vlined]{algorithm2e}
\usepackage{adjustbox}
\usepackage{algorithmicx}
\usepackage{subcaption}
\pdfoutput=1

\definecolor{cvprblue}{rgb}{0.21,0.49,0.74}
\usepackage[pagebackref,breaklinks,colorlinks,allcolors=cvprblue]{hyperref}

\title{\textit{ArcSin: Adaptive ranged cosine Similarity injected noise }\\for Language-Driven Visual Tasks}

\author{
    Yang Liu$^1$, Xiaomin Yu$^2$,  Gongyu Zhang$^1$, Zhen Zhu$^3$, Christos Bergeles$^1$, \\Prokar Dasgupta$^1$, Alejandro Granados$^1$, Sebastien Ourselin$^1$
    \\[2mm]
    $^1$King's College London\ \ 
    $^2$Xreal\ \
    $^3$University of Illinois Urbana-Champaign
    \\
   {\tt\small \{yang.9.liu,firstname.secondname\}@kcl.ac.uk,  yuxm02@gmail.com, zhenzhu4@illinois.edu}
}

\begin{document}
\maketitle
\def\method{\textit{ArcSin}}
\begin{abstract}


\textbf{``A data scientist is tasked with developing a low-cost surgical VQA system for a 2-month workshop. Due to data sensitivity, she collects 50 hours of surgical video from a hospital, requiring two months for privacy approvals. Privacy restrictions prevent uploading data to platforms like ChatGPT, so she assembles one annotator and a medical expert to manually create QA pairs. This process takes three weeks and costs over \$10,000. The trained model provides accurate responses within the limited data scope but lacks broader generalizability, completing the project in 3 months."}

To simplify the challenges presented in the scenario above. In this paper, we replace the image input with text for Vision-language training. Inspired by prior noise injection methods to reduce modality gaps, we introduce Adaptive ranged cosine Similarity injected noise (ArcSin). First, we introduce an innovative adaptive noise scale that effectively generates the textual elements with more variability while preserving the original text feature's integrity. Second, a similarity pool strategy is employed, expanding the domain generalization potential by broadening the overall noise scale. This dual strategy effectively broadens the scope of the original domain while safeguarding content integrity. Our empirical results demonstrate that these models closely rival those trained on images in terms of performance. Specifically, our method exhibits substantial improvements over the previous state-of-the-art, achieving gains of 1.9 and 1.1 CIDEr points in S-Cap and M-Cap, respectively. Additionally, we observe increases of 0.5 percentage points (pp), 1.4 pp, and 1.4 pp in accuracy for VQA, VQA-E, and VE, respectively, pushing the boundaries of what is achievable within the constraints of image-trained model benchmarks.

\end{abstract}

\section{Introduction}
\label{sec:intro}


Training vision models often requires large image datasets and extensive manual annotation, making the process costly and time-consuming. Language-driven visual training offers a ``free lunch" alternative, leveraging text-based scenarios generated by domain experts or large language models (\eg, ChatGPT~\cite{chatgpt}, LLaMA~\cite{touvron2023llamaopenefficientfoundation}). By simulating visual training without image-label pairs, this approach significantly reduces costs and is ideal for applications with moderate accuracy requirements, such as short-term needs, auxiliary tools, or preliminary screening tasks.

Despite its advantages, a key challenge remains: bridging the domain gap between image and text representations, which limits performance when relying solely on text-driven training. Nevertheless, language-driven methods provide an efficient and scalable solution for scenarios where low cost and speed outweigh the need for high precision.


\begin{figure}[t!]
\centering
\includegraphics[width=0.8\linewidth]{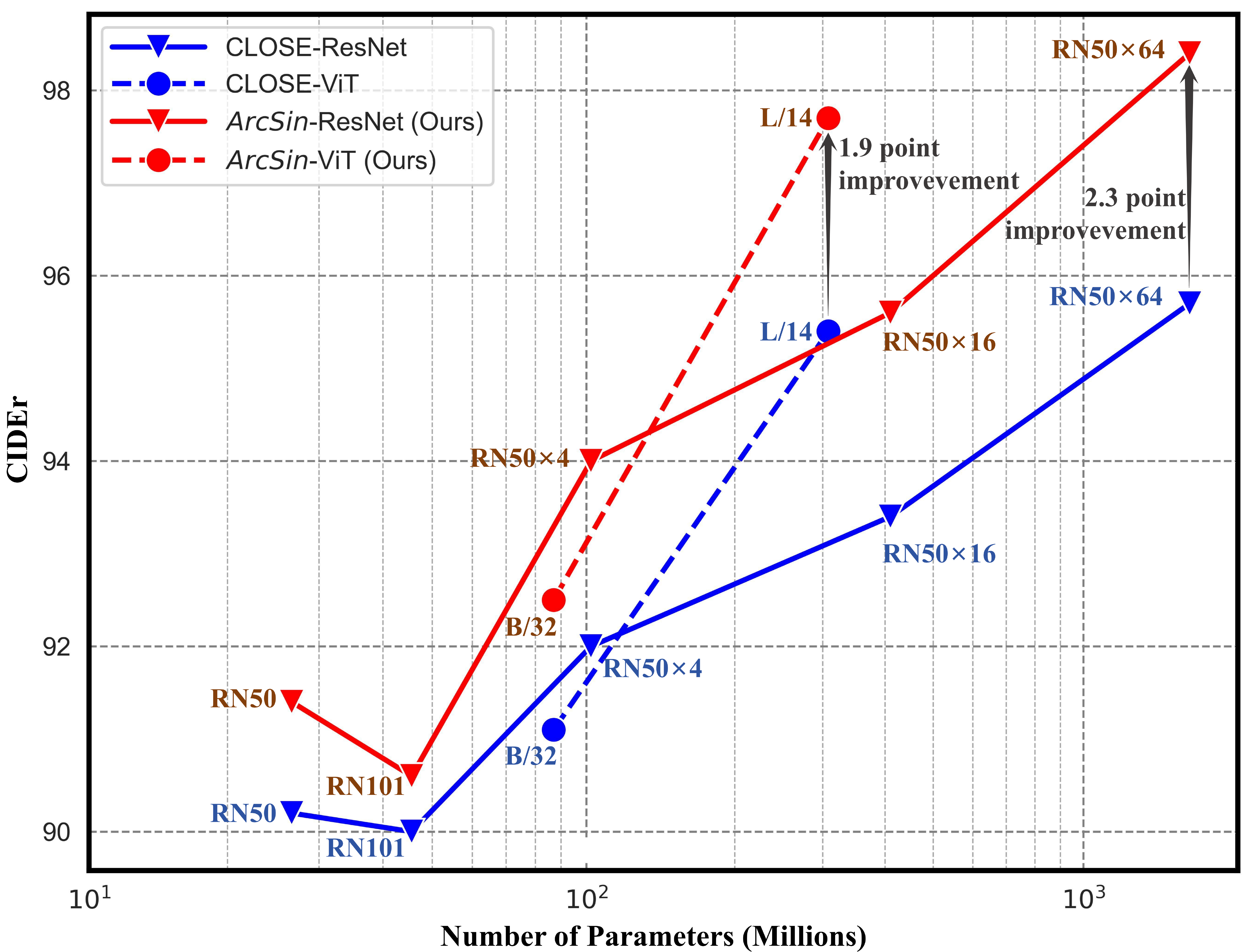}
\caption{\textbf{\method{}~surpasses state-of-the-art method CLOSE~\cite{gu2022can} in the Single Image Captioning (S-Cap) task while utilizing various CLIP models~\cite{Radford2021-CLIP} as contrastive backbones for image-text feature alignment and maintaining a consistent language model (T5-base~\cite{raffel2020exploring}).} The graph illustrates the relationship between the number of model parameters and corresponding CIDEr~\cite{vedantam2015cider} scores.}
\label{fig:intro_different_bb}
\centering
\vspace{-4mm}
\end{figure}

The usual approach to training and extending vision-language models involves using a contrastive loss \cite{Radford2021-CLIP, Ramesh2021-DALL-E, Jia2021-ScalingUp}, which reduces the gap between image and text modalities. Mitigating the modality gap is challenging for multimodal tasks, such as cross-modal retrieval and multimodal fusion~\cite{Wang2021-Introduction, Zhang2022-Multimodal}. 
Previous studies have explored various methods to bridge this modality gap. For example, approaches like VilBERT \cite{Lu2019-VilBERT} and UNITER \cite{Chen2020-UNITER} have demonstrated considerable results in aligning visual and textual representations through joint training strategies. Still, those methods encounter the challenge of overfitting due to the persistent domain gap between language and vision modalities.

One popular approach to reduce overfitting is noise injection. However, the application of noise injection to close the modality gap presents a critical challenge: a trade-off between maintaining the integrity of the original embedding and bridging the gap between modalities. Our research tackles this problem by carefully controlling how much noise we add, ensuring both the narrowing of the modality gap and the preservation of semantic fidelity. We built the method upon studies such as \cite{Gong2021-yb,Goodfellow2014-xy,Xie2020-NoiseInjection}, which highlight the role of noise in strengthening model resilience. Recently, methods in works \cite{gu2022can, Huang2021-eaan} have applied noise injection to improve cross-modal transfer, promoting generalization across varied modalities pre-aligned by contrastive loss.

We present a novel adaptive noise technique known as \textit{Adaptive ranged cosine Similarity injected noise} (\textbf{\method{}}). \method{} optimizes noise injection based on feature values, offering a more adaptable approach that aligns with the fundamental nature of features shaped by contrastive loss, which aims to maintain high similarity scores across different domains with comparable semantic content. 
To prove the effectiveness of our method, we have designed experiments on various language-driven vision-language tasks, which are exclusively trained on textual descriptions and utilize image inputs only during testing. This domain is economically viable and is rich in data availability, given the large repositories of text on the Internet. The vision-language tasks include Visual Question Answering (VQA), Image Captioning (IC), and Visual Entailment (VE). As depicted in Fig.~\ref{fig:intro_different_bb}, our approach achieves enhanced performance across a range of contrastive models tasked with aligned text-image features.

Our contributions are threefold. First, we introduce an adaptive noise injection technique that is tuned based on a similarity threshold and a feature magnitude that ensures content integrity while diversifying the feature space to enable better domain generalization.
Second, we present an injection pool strategy that broadens the amplitude of the injected noise without affecting the similarity with respect to the original features. This strategy effectively improves the cross-domain generalization.
Third, our method improves the capabilities of existing contrastive-based models, such as CLIP~\cite{Radford2021-CLIP}, by establishing a new framework for cross-modal transfer. We demonstrate that our method successfully bridges the modality gap via extensive experiments, beating existing methods in language-driven visual tasks.
\section{Related Work}
\label{sec:relatedwork}

\noindent \textbf{Bridging the Modality Gap in Multi-modal Learning.}
Contrastive loss-driven Vision and Language models (VLMs) aim to minimize the modality gap by closely aligning text and image vectors, ensuring matching pairs are near while unrelated ones are separated \cite{Liang2022-ga}. This gap results from the specific dynamics of model initialization and contrastive optimization \cite{Liang2022-ga}. Jiang~\etal \cite{Jiang2023-pa} and Qian~\etal \cite{Qian2023-xu} showed perfect alignment might not always benefit downstream tasks, suggesting the creation of meaningful latent spaces and emphasizing label relationships. Udandarao~\etal \cite{Udandarao2020-mf} explored class dynamics within these spaces. Technologies like Flamingo \cite{alayrac2022flamingo} and Blip-2 \cite{li2023blip} have pushed the boundaries with alignment techniques that increase the efficacy of cross-attention layers in integrating visual cues into language models. Wang~\etal \cite{Wang2023-lb} proposed the C-MCR, a method for semantic consistency across modalities.

In language-driven vision models, especially for image captioning, recent strides have been made towards training-free methods, enhancing visual understanding directly from textual input. Tewel~\etal \cite{tewel2022zerocap} combined CLIP with a language model, leveraging CLIP’s visual proficiency without additional training. Similarly, Su~\etal \cite{su2022language} introduced visual controls into language generation, enabling zero-shot capabilities. Conversely, Wang~\etal \cite{wang2023association} relied on a rich corpus of captions for training consistency, stressing the dependency on data quality. ICSD~\cite{ma2023text} and SynTIC \cite{liu2023improving} integrate a pre-trained text-to-image (t2i) model for better feature synchronization but at the cost of increased prior knowledge and narrower scope.


\noindent \textbf{Enhancing Robustness Through Noise Injection.}
Noise injection has been crucial in improving model robustness and domain generalization, as evidenced by Wang~\etal~\cite{wang2019detecting} in adversarial defense and Kaji and Kida \cite{kaji2019overview} in medical imaging enhancements. Ji~\etal's~\cite{Ji2020-pz} work on making synthetic images more realistic and Zhen~\etal~\cite{zhen2020flow} in brain imaging further exemplify the utility of noise across different fields. In vision-language integration, leveraging noise to learn visual tasks through language supervision has challenged conventional approaches. For instance, \cite{Kedzie2019-dh} showcases noise's role in neural language generation, while Chen~\etal \cite{Chen2023-bu} apply it in vision-language pre-training for zero-shot recognition. Notably, Nukrai~\etal \cite{nukrai2022text}, Gu~\etal \cite{gu2022can}, and Li~\etal \cite{li2023decap} explore the noise injection of text feature injection for enhanced cross-modal inference, though their methods lack control, potentially altering content.
\section{Problem Formulation}

\begin{figure*}
\centering
\includegraphics[width=0.85\linewidth]{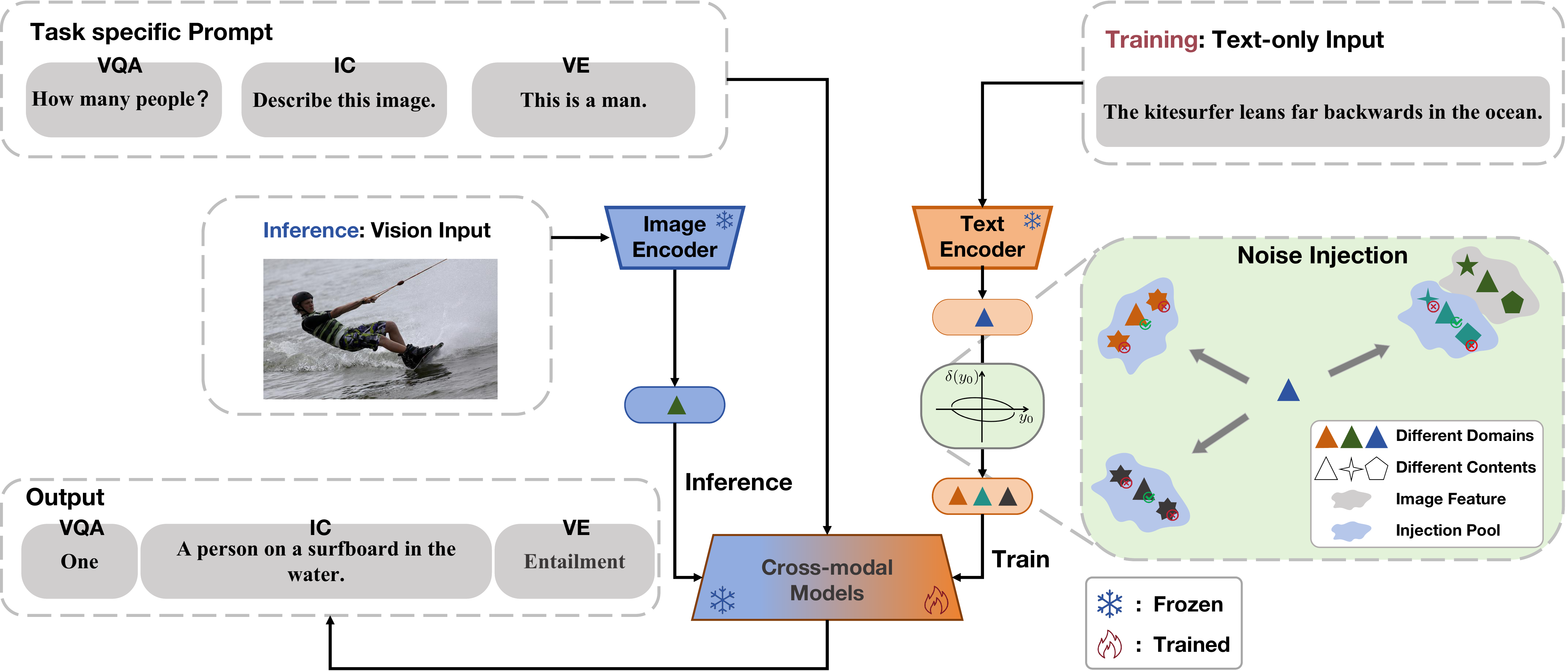}
\caption{\textbf{The \method{}~Architecture.} In the training phase, textual descriptions are encoded to feature vectors using a pre-trained text encoder. These vectors are then augmented with dynamically injected noise, aligning them with vision-language tasks through task-specific prompts within the cross-modal models, such as the T5 model~\cite{raffel2020exploring}. During inference, images are processed into feature vectors via an image encoder, which then replaces the text-derived features, allowing the model to perform visual tasks using text-trained embeddings.}
\label{fig:overview}
\centering
\end{figure*}

In this work, we explore an interesting task that aims to improve the performance of text-driven vision models using text-only training data. A visual task can be defined as a function that maps the visual input $x_v$ to a desired output $y$, prompted by a task-specific directive $q$, such as the query in Visual Question Answering (VQA) scenarios and the null value in Image Captioning (IC) contexts. In training, we replace training images $x_v$ with their text captions $x_t$; the inference procedure still uses visual images as input. This training and testing paradigm assumes that text and image features are closely related. Vision-Language models (VLMs) like CLIP~\cite{Radford2021-CLIP}, with their dual-encoders, text encoder $E_t$ and image encoder $E_v$, have shown that aligning text and image features through contrastive loss can be effective. We use text embeddings $e_t = E_t(x_t)$ for training and image embeddings $e_v = E_v(x_v)$ for inference, ensuring that the model can interpret images based on what it has learned from text. The overview of our method is shown in Fig.~\ref{fig:overview}, where text and image encoders are frozen and cross-modal models, like T5 model~\cite{raffel2020exploring}, are learnable.

\section{Adaptive ranged cosine Similarity injected noise (\method{})}

\label{sec: our_method}
\subsection{Adaptive Ranged Noise}
\begin{figure}[t]
    \centering
    \begin{subfigure}[t]{0.12\textwidth}
        \centering
        \includegraphics[width=\linewidth]{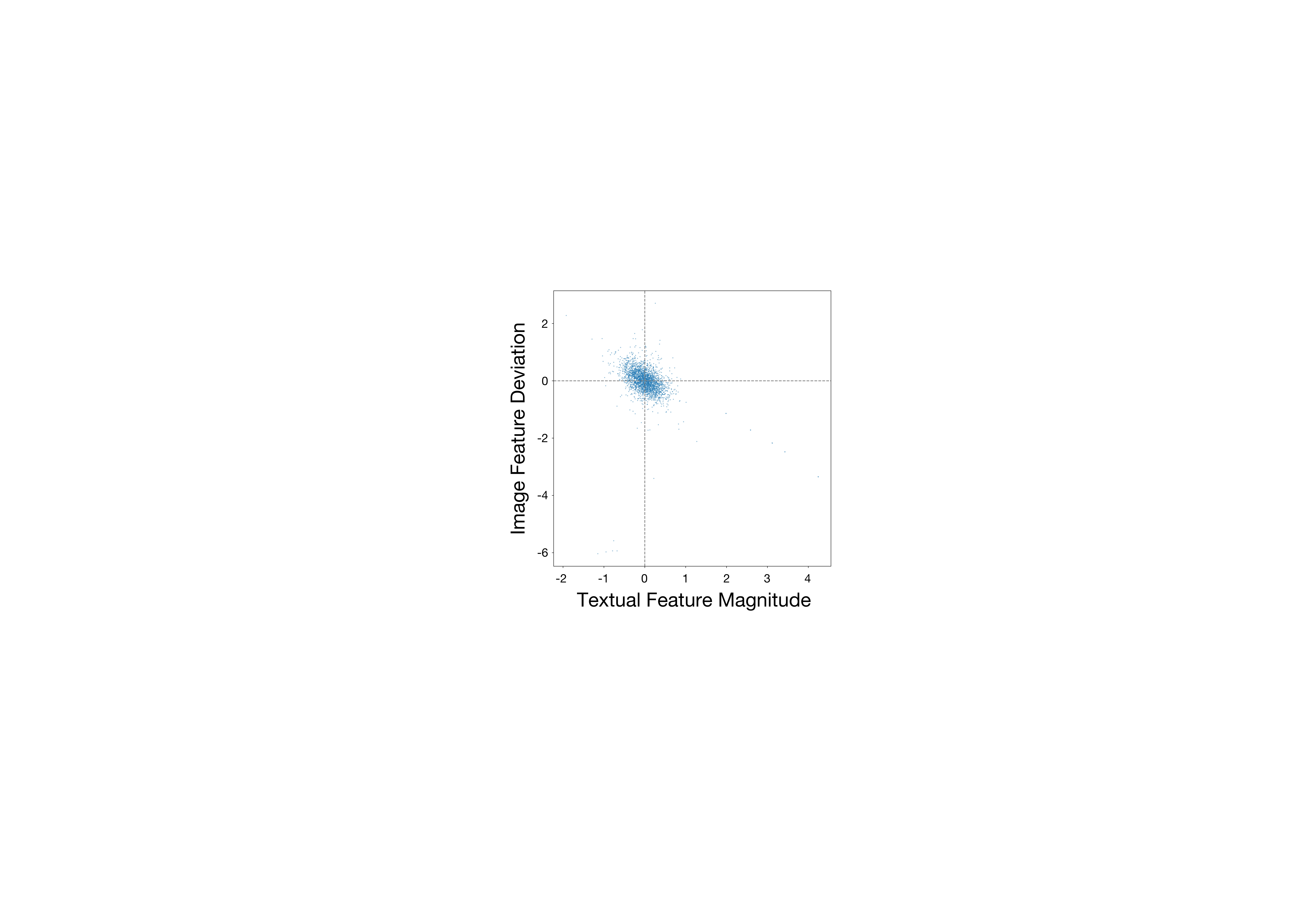}
        \caption{}
        \label{fig:visulize_textimage_change}
    \end{subfigure}%
    \hspace{2mm} 
    \begin{subfigure}[t]{0.33\textwidth}
        \centering
        \includegraphics[width=\linewidth]{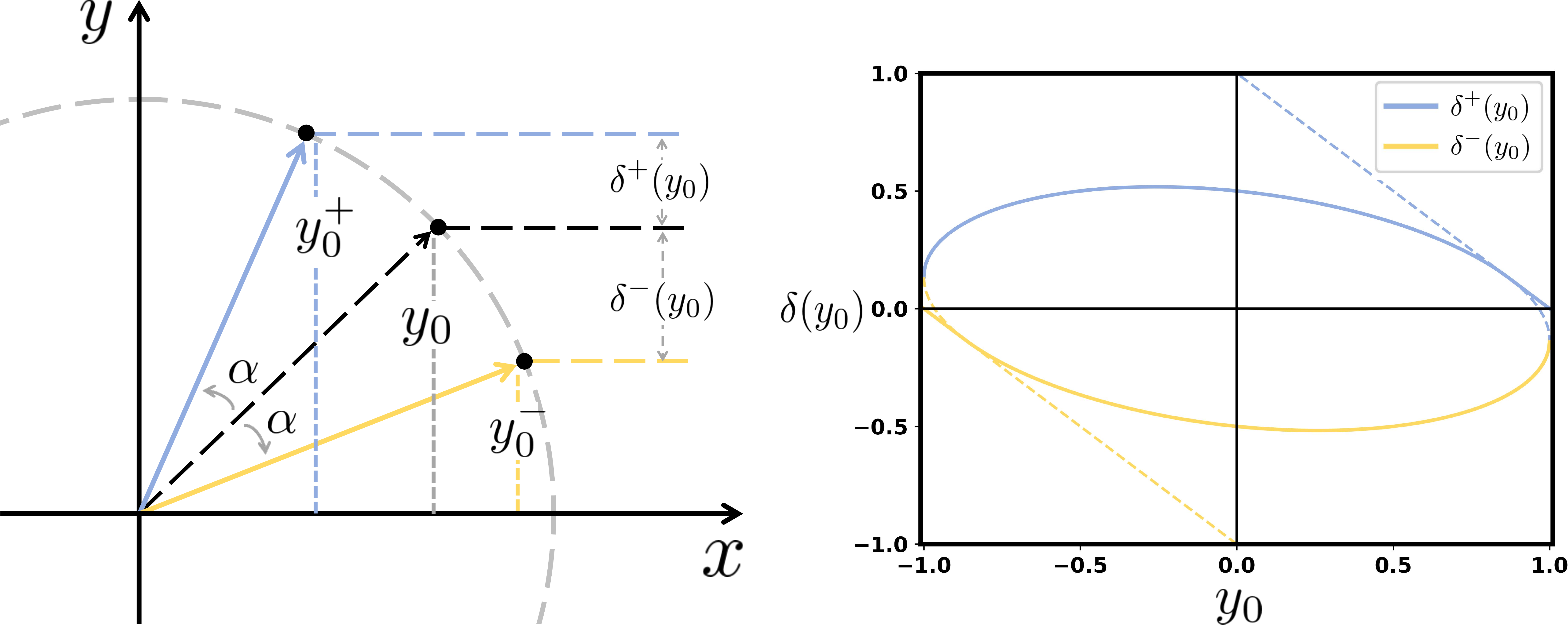}
        \caption{}
        \label{fig:formula}
    \end{subfigure}
    \caption{\textbf{The visualization of feature alignment in cosine similarity controlled multimodal data processing.} \textbf{(a).} Illustration of the relationship between text feature value and corresponding deviations of text-image features. We randomly selected 5 text-image feature pairs encoded by CLIP (ViT-B/32)~\cite{Radford2021-CLIP} and visualized a set of $512 \times 5$ points depicting value deviations. It plots the text feature values along the horizontal axis against the corresponding deviations in the image features on the vertical axis. It indicates variable deviations across different magnitudes of textual features. \textbf{(b).} The visualization of permissible value deviations under a predefined similarity threshold in standard 2D space. \textbf{Left:} A vector (in black) with a vertical component $y_0$. Ensuring that the cosine similarity between the noise-augmented vectors (in blue and yellow) and the original vector remains above a certain threshold equates to confining rotation within an angle $\alpha$. The allowable positive and negative deviations are denoted by $\delta^+(y_0)$ and $\delta^-(y_0)$, respectively. \textbf{Right:} The relationship between $y_0$ and its potential deviations $\delta(y_0)$.}
    \label{fig:both_images}
\vspace{-2mm}
\end{figure}

Training with a mini-batch of encoded text features \(e_t \in \mathbb{R}^{B \times C}\), where $B$ and $C$ represent the batch size and feature vector length, respectively, we are faced with the challenge of a lack of actual visual data.
Accordingly, our aim extends beyond merely bridging domain gaps; we seek to strengthen the model's capacity to understand visual content via text-based training, thereby countering overfitting. Adding Gaussian noise to the text feature, as suggested by~\cite{gu2022can}, is a straightforward approach.
However,
it limits itself to uniformly applying the same intensity of noise across all feature elements. Such a blanket strategy is ineffective because the content features are prone to be significantly changed or even destroyed during noise injection. Fig.~\ref{fig:visulize_textimage_change} brings to light the variance between text feature values and the resulting deviations in corresponding image features, indicating that a more discerning approach is needed. Our proposed method injects noise in an adaptive manner, taking into account the individual magnitude of text features to ensure domain generalization is achieved with less content damage.

To analyse the cosine similarity resulting from a contrastive loss training regime more easily, let's consider a two-dimensional scenario. Imagine a 2D vector $\vec{v_0} = (x_0, y_0)$ subject to a rotation by an angle $\alpha$, which represents noise introduction. We need to ensure that this rotated vector stays within a threshold of similarity to the original vector. For every component in $\vec{v_0}$, like $y_0$, the permissible adjustment is constrained, as $\delta(y_0)$, to ensure the inherent meaning of the data is not distorted.

As detailed in the left of Fig.~\ref{fig:formula}, we derive bounds for positive and negative adjustments, $\delta^+ (y_0)$ and $\delta^- (y_0)$, which are defined as follows:
\begin{small}
\begin{align}
\delta^+ (y_0) &= \begin{cases} 
\sin(\arcsin(y_0) + \alpha) - y_0, &  \arcsin(y_0) + \alpha < \frac{\pi}{2} \\
1 - y_0, & \text{otherwise.}
\label{eq:delta_positive}
\end{cases}  \\
\delta^- (y_0) &= \begin{cases} 
\sin(\arcsin(y_0) - \alpha) - y_0, &  \arcsin(y_0) - \alpha > -\frac{\pi}{2} \\
-1 - y_0, & \text{otherwise.}
\label{eq:delta_negative}
\end{cases}
\end{align}
\end{small}

These equations inform us of the range within which $y_0$ can be altered without significantly changing the content, and the curves are shown to the right of Fig.~\ref{fig:formula}, which illustrate the varying degrees of permissible noise depending on the initial value of $y_0$.

Extending this principle back to high-dimensional feature space, the noise augmentation process for each feature dimension $j$ is determined by its original text feature value $e^j_t$, following the mechanism:
\begin{align}
\hat{e}^j_t &= \begin{cases}
e^j_t + \delta^+(e^j_t) \cdot \xi, & \quad  \xi \geq 0, \\
 e^j_t - \delta^-(e^j_t) \cdot \xi, & \quad otherwise,
\end{cases}
\end{align}
 where $\delta^+$ and $\delta^-$ are scaling functions derived from the bounds established by 2D Eqs.~\ref{eq:delta_positive} and~\ref{eq:delta_negative}, and $\xi \sim \mathcal{N}(0, 1)$ denotes the standard Gaussian distribution. Within the defined range of the $\textit{arcsin}$ function, we use a clamping operation to ensure that the input values are kept within the interval [-1, 1] as the feature is already normalized by its L2 norm. (Algo.~\ref{algorithm_arcsin}, line 5).

Our adaptive noise injection starts from cosine similarity to adeptly generalize across new domains for multimodal features. It strikes a balance between domain expansion and content preservation.

\newcommand{\smalltcp}[1]{\tcp{\fontsize{6pt}{8pt}\selectfont #1}}
\begin{algorithm}[t]
\caption{Adaptive Ranged Cosine Similarity Injected Noise (\method{}) (Sec.~\ref{sec: our_method})}
\label{algorithm_arcsin}
\KwIn{Encoded text features $e_t \in \mathbb{R}^{B \times C}$, similarity threshold $s$, noise pool size $N_p$}
\KwOut{Noise-augmented text features $\hat{e}_t$}
\SetKwProg{Class}{Class}{:}{end}
\SetKwProg{Init}{def \_\_init\_\_}{:}{end}
\SetKwProg{Forward}{def forward}{:}{end}
\SetKwFunction{FMain}{ArcSin}
\SetKwFunction{FInit}{\_\_init\_\_}
\SetKwFunction{FForward}{forward}

\Class{\FMain{}}{
    \Init{($s, N_p$)}{
        \smalltcp{Set initial angle range, similarity threshold, and noise pool size}
        self.$\alpha_r \gets [0, \frac{\pi}{2}]$; self.$s \gets s$; self.$N_p \gets N_p$;

    }
    \Forward{($e_t$)}{
        $e_t \gets e_t.clamp(-1, 1)$; \smalltcp{Clamp features}
        $\alpha_0 \gets$ Random(self.$\alpha_r$); \smalltcp{Select random angle}
        
        \smalltcp{Dynamic scale calculation for noise}
        $\delta^+(e_t) \gets (e_t + \alpha_0 < \frac{\pi}{2}) ? \sin(\arcsin(e_t) + \alpha_0) - e_t : 1 - e_t$;
        
        $\delta^-(e_t) \gets (e_t - \alpha_0 > -\frac{\pi}{2}) ? \sin(\arcsin(e_t) - \alpha_0) - e_t : -1 - e_t$;
        
        \smalltcp{Generate noise pool and augment features}
        $\mathcal{P}_n \gets \text{GenerateNoisePool}(B, \text{self.}N_p, C)$; 
        
        $\hat{\mathcal{E}} \gets e_t + \mathcal{P}_n \times ( (\mathcal{P}_n > 0) ? \delta^+(e_t) : (-\delta^-(e_t)) $;

        $\hat{e}_t \gets \arg\max \text{Sim}(e_t, \hat{\mathcal{E}})$; \smalltcp{Select best noise vector based on similarity}
        
        \smalltcp{Update angle range based on similarity}
        \uIf {$\text{AverageSim}(e_t, \hat{e}_t) < s - \epsilon$}{
            self.$\alpha_r.\text{upper} \gets \alpha_0$; 
        }\uElseIf {$\text{AverageSim}(e_t, \hat{e}_t) > s + \epsilon$}{
            self.$\alpha_r.\text{lower} \gets \alpha_0$; 
        }
        
        \KwRet $\hat{e}_t$; \smalltcp{Return augmented feature}
    }
}
\vspace{-3mm}
\end{algorithm}

\subsection{Injection Pool}
To balance content preservation and the variability by injected noise, we carefully calibrate the noise injection scale. If we apply noise conservatively, we risk limiting the extent of domain generalization, potentially weakening the model's ability to perform well across diverse domains. To address this, we introduce a novel, yet straightforward technique: creating a noise injection pool.

As shown in Algo.~\ref{algorithm_arcsin} (line 9-10), for each encoded text feature vector $e_t$, we generate $N_p$ potential noise vectors. These vectors constitute our injection pool, a diverse set of candidates for domain generalization. We then experimentally add each noise vector to the original feature, computing the cosine similarity between the original $e_t$ and noise-augmented version. The noise vector that yields the highest similarity is selected to create the final injected feature $\hat{e_t}$.

This method achieves two vital objectives: increase the magnitude of domain variation while maintaining the similarity score between the original and noise-injected features. Consequently, this elevates the likelihood of synthesizing features representative of a broader spectrum of domains, enhancing the model's domain generalization capabilities.

\subsection{Similarity Threshold Controlled Injection}

To gain better control over the noise injection process, we employ a strategy governed by a predefined similarity threshold $s$. This threshold dictates the acceptable bounds of deviation from the original feature space during noise application. In practice, during training, we define an initial rotation angle \(\alpha\) within the interval \([0, \frac{\pi}{2}]\) (Algo.~\ref{algorithm_arcsin}, line 3), corresponding to the potential alteration in feature space.

Upon the arrival of a new mini-batch feature $e_t$, we select a random angle $\alpha_0$ from within the predefined range. This angle serves as a pivot point to assess the average cosine similarity between the original feature $e_t$ and the noise-enhanced feature $\hat{e_t}$. Based on the similarity score, As depicted in Algo.~\ref{algorithm_arcsin} (line 12-15), we dynamically adjust the angle bounds:

\begin{itemize}
\item If the similarity score falls below $s - \epsilon$, where $\epsilon$ is a small constant, we tighten the upper bound of $\alpha$ to $\alpha_0$.
\item If the similarity score exceeds $s + \epsilon$, we expand the lower bound of $\alpha$ to $\alpha_0$.
\item If the score is within the threshold range, no changes.
\end{itemize}

This iterative adjustment ensures that the noise we introduce stays within a range that preserves content fidelity, as indicated by our similarity metric, while still providing sufficient variability for robust domain generalization.

\section{Experiments}

\begin{table*}[t!]
\centering
\caption{\textbf{Comparison with previous work on different V\&L tasks.} Most of them are proposed for specific tasks, and only CLOSE and our \method{} can process all the provided tasks. We report BLEU-4 and CIDEr scores for S-Cap and M-Cap, and accuracy percentages for VQA, E-VQA and VE. `zero-shot' indicates training on other datasets within images. we take training within images as the upper bound. The highest scores are in bold.}
\begin{subtable}{\textwidth}
\centering
\setlength{\tabcolsep}{6.6pt}
\renewcommand{\arraystretch}{1.2}
\resizebox{0.75\linewidth}{!}{
\begin{tabular}{c|cccc|ccccc|c}
\toprule 
\hline
\multirow{3}{*}{M-Cap} & \multicolumn{4}{c|}{2022} & \multicolumn{5}{c|}{2023} & \multicolumn{1}{c}{2024} \\
\cline{2-11}
& ZeroCap & MAGIC  & ESPER Style  & CapDec  & ICSD  &\multicolumn{1}{c|}{synTIC}& DeCap  & MultiCapCLIP   & Knight & \method{} \\
&\cite{tewel2022zerocap} &\cite{su2022language} &\cite{yu2022multimodal} &\cite{nukrai2022text} &\cite{ma2023text} &\cite{liu2023improving} &\cite{li2023decap} &\cite{yang2023multicapclip} &\cite{wang2023association} & \\
\hline
Text-only & \checkmark & \checkmark & \checkmark & \checkmark &\multicolumn{2}{c|}{\checkmark(Need t2i model)} & \checkmark & \checkmark & \checkmark & \checkmark \\
\hline

\hline
B@4 $\uparrow$ & 7.0 & 12.9 & 21.9 & 26.4 & 29.9 &\multicolumn{1}{c|}{29.9}& 24.7 & 27.6  & 27.8 & \textbf{30.3}\\
C $\uparrow$& 34.5 & 49.3 & 78.2 & 91.8 & 96.6 &\multicolumn{1}{c|}{\textbf{101.1}} & 91.2 & 96.2 & 98.9 & 99.6\\
\midrule 
\bottomrule 
\end{tabular}
}
\vspace{2mm}
\label{subtable:compare_with_sata_1}
\end{subtable}

\begin{subtable}{\textwidth}
\centering
\renewcommand{\arraystretch}{1.2}
\resizebox{0.75\linewidth}{!}{
\begin{tabular}{c|cccc|cc}
\toprule 
\hline
\multirow{2}{*}{Tasks} & \multicolumn{4}{|c}{VQA} & \multicolumn{2}{|c}{VE}  \\
\cline{2-7}
& TAP-C \cite{song2022clip} & BLIP-2 ViT-G \cite{li2023blip} & Flamingo 80B\cite{alayrac2022flamingo} & \textbf{\method{}} & CLIP Cls.\cite{song2022clip}  & \textbf{\method{}} \\
\hline
Text-only & \checkmark & zero-shot & zero-shot & \checkmark & \checkmark & \checkmark   \\
\hline
Acc $\uparrow$ & 38.7 & 52.6 & 56.3 & \textbf{61.8} & 66.6 & \textbf{76.0}\\
\midrule 
\bottomrule 
\end{tabular}
}
\label{subtable:compare_with_sata_2}
\end{subtable}
\vspace{2mm}

\begin{subtable}{\textwidth}
\centering

\renewcommand{\arraystretch}{1.2}
\resizebox{0.75\linewidth}{!}
{
\begin{tabular}{c|c|c|ccccccc} 
\toprule 
\hline
\multirow{2}{*}{Tasks} &\multirow{2}{*}{Model} & \multirow{2}{*}{Text-only} & \multicolumn{2}{c}{S-Cap} & \multicolumn{2}{c}{M-Cap} & VQA & VQA-E & VE \\
\cline{4-10}
&& & B@4$\uparrow$& C $\uparrow$ & B@4 $\uparrow$ & C $\uparrow$ & Acc $\uparrow$ & Acc $\uparrow$ & Acc $\uparrow$\\
\midrule 
\multirow{4}{*}{All}&W/o Noise & \checkmark & 4.2 & 16.4 & 21.9 & 68.7 & 60.2 & 59.8 & 68.2 \\
&CLOSE \cite{gu2022can} & \checkmark & 28.2 & 95.8 & 29.6 & 98.5 & 61.3 & 63.4 & 74.6 \\
&\textbf{\method{~(Ours)}} & \checkmark & \textbf{29.4} & \textbf{97.7} &\textbf{30.3}  &\textbf{99.6}  &\textbf{61.8}  & \textbf{64.8} & \textbf{76.0} \\
\cline{2-10}
&W/Image (Upper bound) &  & 34.4 & 113.2 & 34.4 & 113.2 & 65.4 & 67.9 & 77.7  \\
\hline
\bottomrule 
\end{tabular}
}
\label{subtable:compare_with_sata_3}
\end{subtable}

\label{table:compare_with_sata}
\end{table*}
\subsection{Datasets and Settings}

To demonstrate the effectiveness of our model, we conducted experiments on three vision-language tasks: Image Captioning (IC), Visual Question Answering (VQA), and Visual Entailment (VE). 

\subsubsection{Datasets}

\textbf{Image Captioning (IC).} For the Image Captioning task, we employed the Karpathy split \cite{karpathy2015deep} to partition the COCO Captioning \cite{chen2015microsoft} dataset. During the training process, only the textual data from the training set was utilized to train the text generation model. We defined two training scenarios: Single Captioning (S-Cap) and Multiple Captioning (M-Cap). In Single Captioning, we selected a single text caption as both the input text and the target output text. In M-Cap, considering that an image may have multiple distinct caption descriptions, we used different captions about the same image as input and target output text. Through Multiple Captions, the model was able to learn captions of the same image in various contexts, thereby endowing the generated image captions with a more diverse range of contexts and expressions.

\noindent \textbf{Visual Question Answering (VQA).} For the VQA task, as we cannot access image information, we opt to use sentences containing scene descriptions as substitutes during training. Thus, in the training process, we utilize data that includes scene descriptions, questions, and target answers. We employ two datasets: VQA 2.0 \cite{goyal2017making} and VQA-E \cite{li2018vqa}. In VQA 2.0, we pair image captions from COCO Captioning with questions about the same images, enabling the model to comprehend visual scenes through textual descriptions. To address potential scenarios in which questions in the VQA 2.0 dataset involve details not present in the image captions, we use the VQA-E dataset. This dataset is a subset of VQA 2.0, with captions from COCO Captioning verified to contain answers to the questions. This ensures that during training, questions can be answered solely by text. We evaluated the model separately on the VQA 2.0 test-dev set and the VQA-E validation set. These two settings will be called `VQA' and `VQA-E', respectively.

\noindent \textbf{Visual Entailment (VE).} This task aims to determine whether there is an entailment, contradiction, or neutral relationship between a given image of a premise and a hypothesis sentence. Since images cannot be obtained during training, we choose to use textual captions as replacements. We conducted training using the SNLI~\cite{maccartney2008modeling} dataset and evaluated the model on the SNLI-VE~\cite{goyal2017making} dataset.

\subsubsection{Metrics}

For the IC task, we primarily use CIDEr \cite{vedantam2015cider} and BLEU-4 \cite{papineni2002bleu} as evaluation metrics. CIDEr focuses on assessing the consistency between generated captions and reference captions, while BLEU-4 is employed to measure the n-gram matching degree between generated captions and reference captions. For the VQA task, Accuracy serves as the primary evaluation metric. This metric quantifies the proportion of model-generated answers that match the ground-truth answers, providing a comprehensive understanding of the overall accuracy of the model in answering questions related to a given scene image. For the VE task, Accuracy is also employed as the main evaluation metric. Accuracy is used to gauge the model's precision in correctly determining the relationship between the premise image and the hypothesis sentence, offering a clear assessment of the model's performance.

\subsection{Comparisons with State-of-the-art Methods}
We evaluate the benefit of \method{} when used in conjunction with state-of-the-art models across various visual and language tasks, including M-Cap, S-Cap, VQA, VQA-E, and VE. The comparative results are presented in Table~\ref{table:compare_with_sata}. For CLOSE~\cite{gu2022can}, we used their open source code\footnote{https://github.com/allenai/close/tree/main}, noting that our results varied, with some tasks exceeding and others falling below their reported results.
Our analysis revealed that while most studies focus on specific tasks like M-Cap, \method{} achieved improvements of $2.5$ points in BLEU-4 and $1.1$ points in CIDEr. It also provides results comparable to ICSD~\cite{ma2023text} and SynTIC~\cite{liu2023improving}, which require an additional t2i model. Notably, \method{} surpasses task-specific methods in VQA and VE by considerable margins.
Comparing general task methodologies, particularly with CLOSE, we observe both methods enhance performance by introducing noise to the text feature, indicating a persistent domain gap, despite models like CLIP's efforts to minimize it. Compared to the upper bound of the image-trained model, \method{} shows marked enhancements over CLOSE, with increases of $1.9$ and $1.1$ points in CIDEr for S-Cap and M-Cap, and increases of $0.5$ pp (percentage points), $1.4$ pp, and $1.4$ pp in accuracy for VQA, VQA-E, and VE tasks, respectively.

\begin{figure*}[t]
\centering
\includegraphics[width=1.\linewidth]{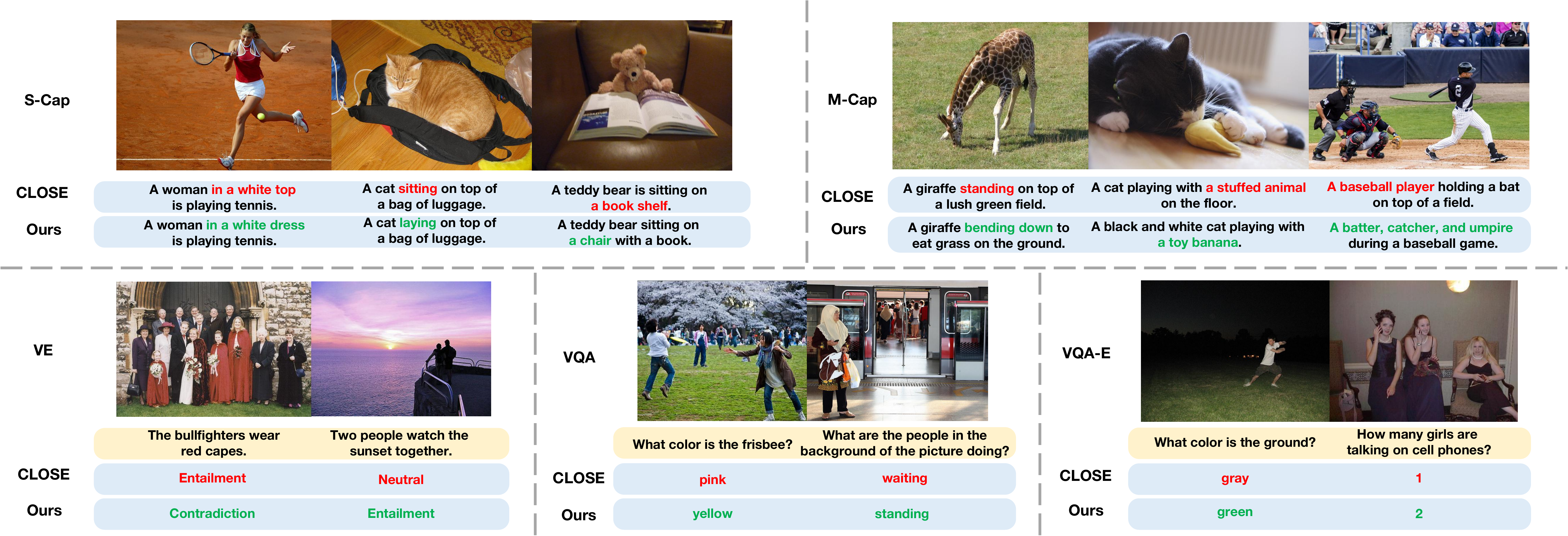}
\caption{\textbf{Qualitative comparisons with the state-of-the-art method CLOSE~\cite{gu2022can}.}}
\label{fig:visulize_sota_correct_examples}
\centering
\vspace{-2mm}
\end{figure*}

Fig.~\ref{fig:visulize_sota_correct_examples} illustrates various qualitative results from multiple tasks, highlighting our method's enhanced capability in interpreting image content. For instance, in the VQA task that asks, `What color is the frisbee?', accurate recognition is challenging here due to the frisbee's small and unclear appearance. Whereas the baseline CLOSE model incorrectly infers `pink', influenced by the predominant cherry blossoms, our method accurately recognizes the frisbee as `yellow'. This highlights our method's adeptness at preserving fine-grained details, which are typically compromised during noise injection. Overall, our \method{} mostly beats CLOSE in distinguishing and retaining nuanced content, proving its superior precision in scenarios where detail accuracy is crucial.

\subsection{Ablation Study}
\noindent \textbf{Performance Analysis of Different Components in \method{}.}~
In light of our method outperforming state-of-the-art approaches across various tasks, we undertook an ablation study to dissect the impact of the individual components within our proposed framework. Tab.~\ref{table:different_components} presents a comparative analysis, illustrating that our adaptive noise injection method yields superior results over the fixed-scale noise approach previously suggested by~\cite{gu2022can}. While the inclusion of an injection pool offers incremental gains, it is the adaptability of our injection technique that stands out as the pivotal factor in enhancing performance. This indicates that while both refining the injection process and expanding the injection space contribute to addressing modality shift challenges, optimizing injection methodology plays a more decisive role.

\begin{table}[ht]
\centering
\caption{\textbf{The results of different components of \method{} on M-Cap, VQA-E, and VE.} BLEU-4 and CIDEr scores are reported on M-Cap, and accuracy percentage is reported on VQA-E and VE.}
\renewcommand{\arraystretch}{1.2}
\resizebox{\linewidth}{!}
{
\begin{tabular}{c|cc|c|c} 
\toprule 
\hline
\multirow{2}{*}{Method} & \multicolumn{2}{c|}{M-Cap} & VQA-E & VE \\
\cline{2-5}
& B@4 & C & Acc & Acc\\
\hline
Fixed Scale Noise (CLOSE) & 29.6  &98.5   & 63.4  & 74.6  \\
Adaptive Ranged Injection &30.0   &99.3   &\textbf{64.8}   & 75.5  \\
Adaptive Ranged Injection + Injection Pool (Ours)& \textbf{30.3}& \textbf{99.6} &\textbf{64.8} & \textbf{76.0}\\
\midrule 
\bottomrule 
\end{tabular}
}

\label{table:different_components}
\end{table}

\begin{table}[ht]
\centering
\caption{\textbf{Comparative Analysis of Vision-Driven Language Tasks S-TCap, TQA-E, and TE.} Performance metrics include CIDEr scores for S-TCap and accuracy percentages for E-TQA and TE.}
\renewcommand{\arraystretch}{1.2}
\resizebox{0.8\linewidth}{!}
{
\begin{tabular}{cccc} 
\toprule 
\hline
Method & S-TCap  &TQA-E & TE \\
\hline
W/o Noise &104.1   &68.6   & 75.4  \\
CLOSE &104.2  &\textbf{69.4}   & 76.6   \\
\method{} (Ours)&\textbf{105.2}  &69.3   & \textbf{77.8}  \\
\midrule 
\bottomrule 
\end{tabular}
}

\label{table:switch_training}
\end{table}

\begin{figure}[t!]
    \centering
    \includegraphics[width=\linewidth]{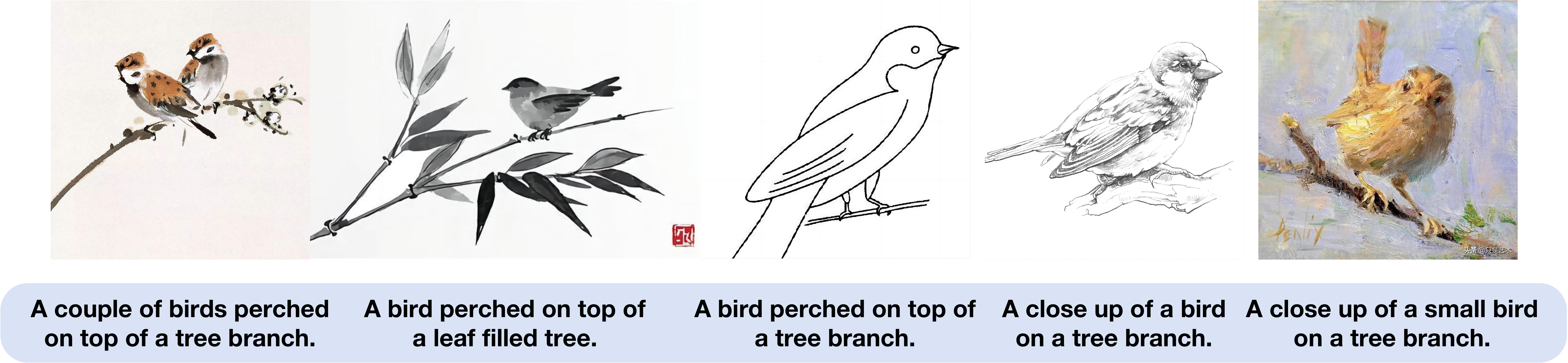}
    \caption{\textbf{Cross-style captioning cases.} Our \method{} model, trained on M-Cap, demonstrates its robust captioning ability on some `bird' images with distinct visual styles, randomly sourced from the web.}
    \label{fig:diff_style_bird}
\end{figure}

\begin{figure}[t!]
    \centering
        \includegraphics[width=0.8\linewidth]{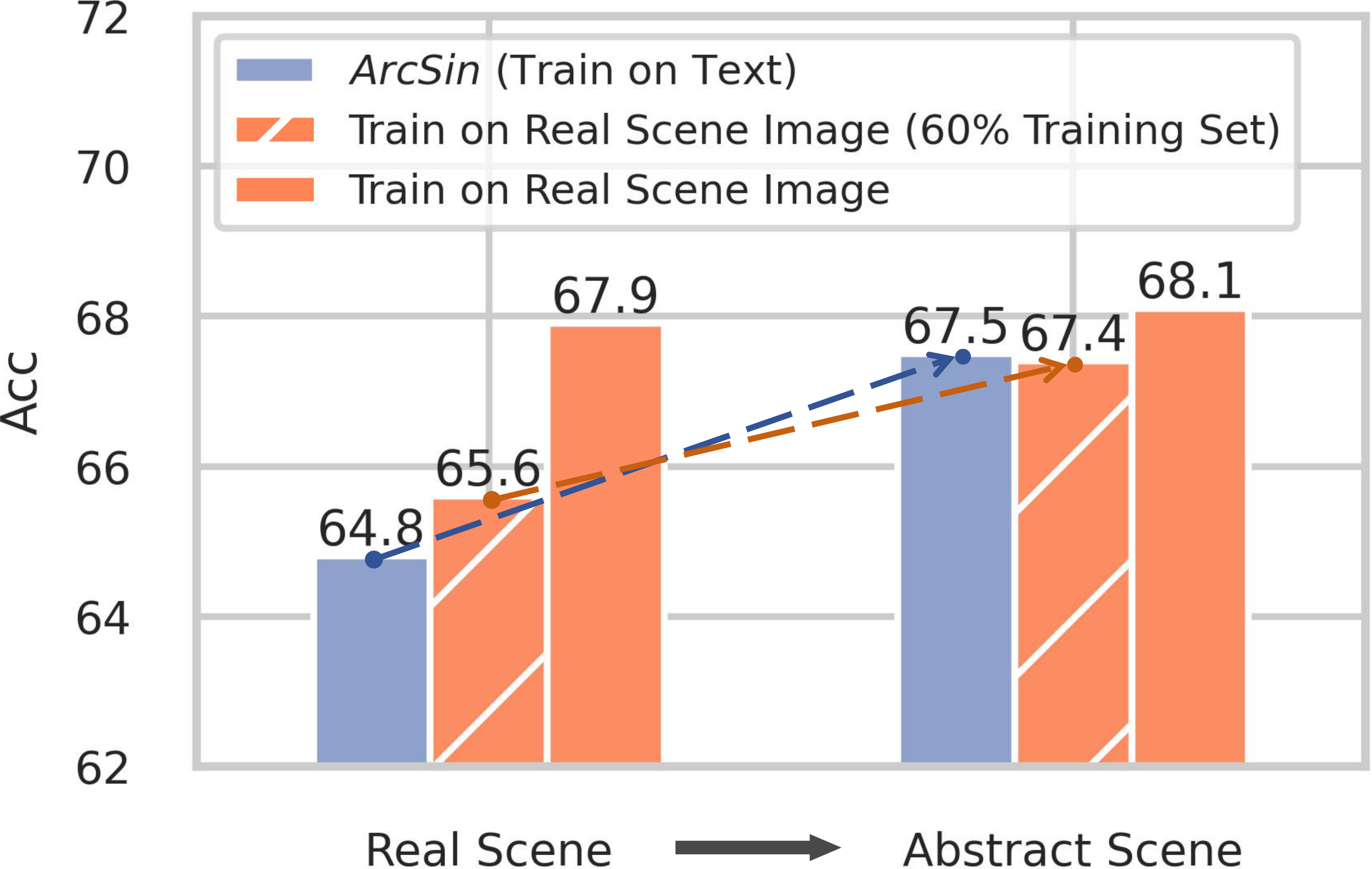}
            \caption{\textbf{Evaluating robustness from real to abstract scenes.} We compare the performance of \method{} (trained on text) with models trained on real-scene images, applied to both real and abstract scenes for the VQA-E task. Additionally, a model trained with 60\% of the image dataset serves as a benchmark to assess the robustness of our method.}
            \label{fig:bar_real_abstract}
\end{figure}

\begin{figure}[t!]
    \centering
        \includegraphics[width=0.8\linewidth]{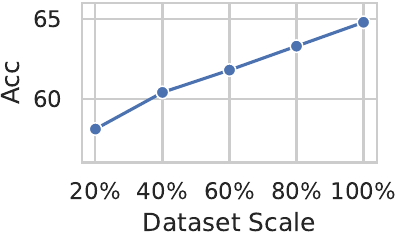}
            \caption{\textbf{Influence of Dataset Scale.} It illustrates the impact of varying dataset sizes on the accuracy of the \method{} model for the VQA-E task, demonstrating a clear trend of improved performance as the amount of training data increases.}
            \label{fig:data_scale}
\end{figure}

\begin{figure*}[t!]
\centering
\includegraphics[width=0.9\linewidth]{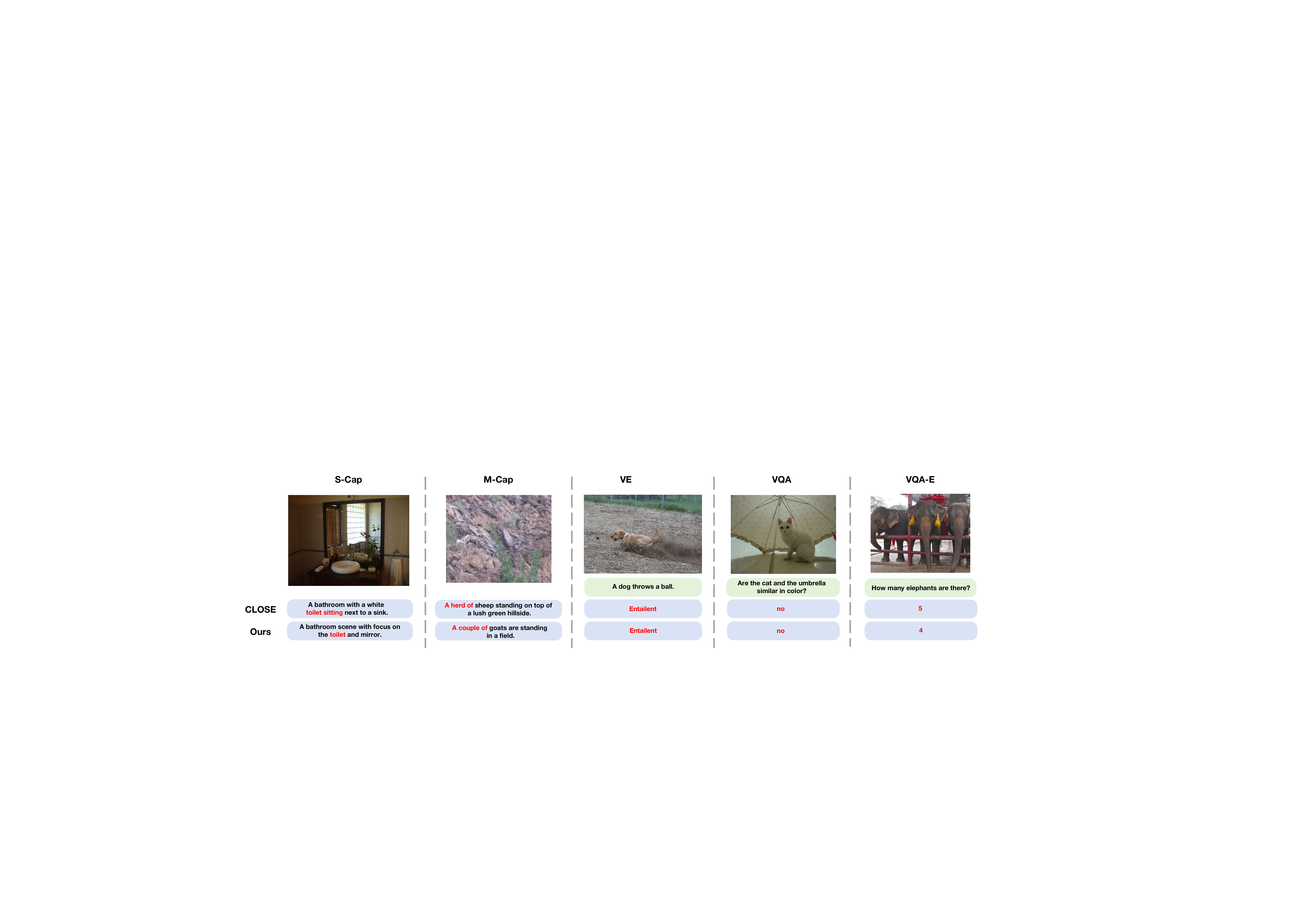}
\caption{\textbf{Some failure cases.}}
\label{fig:fail_cases}
\centering
\vspace{-3mm}
\end{figure*}
\noindent \textbf{Adaptability Across Image Styles.} We've proven \method{}'s effectiveness with photos that capture real life. We now expand our investigation to include a variety of artistic expressions. In Fig.~\ref{fig:bar_real_abstract}, \method{} shows promising results with abstract images from the COCO VQA dataset, hinting at its potential for simpler, stylized visuals. Compared to the model trained with real images, \method{} yields similar results in abstract scenes and even beats the model trained with 60\% of the image data while it trails by a narrow 1 pp margin in real scene evaluation. This contrast highlights \method{}'s adeptness at processing images from different visual domains. Fig.~\ref{fig:diff_style_bird} expands this examination, presenting \method{}'s capabilities in captioning across diverse art styles, further evidencing its robustness and versatility.

\begin{table}[ht]
\centering
\caption{\textbf{Performance comparison of various contrastive and language backbone models utilizing \method{}.} The table outlines the contrastive models and language models in the first two rows, respectively, while the subsequent rows report the BLEU-4 and CIDEr scores for S-Cap and accuracy percentages for VQA-E. The value within parenthesis indicates the performance differential relative to CLOSE~\cite{gu2022can}, with green upward arrow ($\uparrow$)/red downward arrow ($\downarrow$) denoting improvement/reduction.}
\renewcommand{\arraystretch}{1.2}
\resizebox{0.98\linewidth}{!}{
\begin{tabular}{c|ccccccccc}
\toprule
\midrule 
\textbf{CLIP Model} & ViT-L/14 & ViT-L/14 & ViT-L/14 & ViT-B/32 & RN101 & RN50 & RN50$\times$4 & RN50$\times$16 & RN50$\times$64 \\ 
\textbf{T5 Model} & small & base & large & base & base & base & base & base & base \\
\midrule
\textbf{S-Cap B@4} & 29.0 & 29.4 & 28.2 & 27.9 & 27.2 & 27.9 & 28.1 & 28.3 & \textbf{29.7} \\
\textbf{S-Cap C} & 95.7 (\textcolor{green}{1.3$\uparrow$}) & 97.7 (\textcolor{green}{1.9$\uparrow$}) & 96.4 (\textcolor{green}{2.5$\uparrow$}) & 92.5 (\textcolor{green}{1.4$\uparrow$}) & 90.6 (\textcolor{green}{0.6$\uparrow$}) & 91.4 (\textcolor{green}{1.2$\uparrow$}) & 94.0 (\textcolor{green}{2.0$\uparrow$}) & 95.6 (\textcolor{green}{2.2$\uparrow$}) & \textbf{98.4} (\textcolor{green}{2.3$\uparrow$}) \\
\midrule
\textbf{VQA-E Acc} & 63.3 (\textcolor{green}{4.4$\uparrow$}) & 64.8 (\textcolor{green}{1.4$\uparrow$}) & 64.5 (\textcolor{red}{0.7$\downarrow$}) & 62.4 (\textcolor{green}{1.0$\uparrow$}) & 61.7 (\textcolor{green}{1.9$\uparrow$}) & 61.5 (\textcolor{green}{1.1$\uparrow$}) & 63.0 (\textcolor{green}{1.5$\uparrow$}) & 63.6 (\textcolor{green}{1.1$\uparrow$}) & \textbf{65.3} (\textcolor{green}{1.1$\uparrow$}) \\
\midrule 
\bottomrule
\end{tabular}
}
\label{tab:different_bb}
\vspace{-1mm}
\end{table}

\noindent \textbf{The Effects of Different Backbones.}~To establish the impact of different model architectures and validate the strength of \method{}, we conducted a comprehensive evaluation of several CLIP and T5 models. The results, as shown in Tab.~\ref{tab:different_bb}, indicate that when the T5-base model is uniformly applied, the more advanced contrastive backbones ranging from RN50 to RN50$\times$64 consistently deliver superior performance across various tasks. This trend underscores the critical role of effective image-text alignment in improving overall model performance.

Meanwhile, the T5-base model outperformed the T5-large model when paired with the ViT-L/14 CLIP model, suggesting that an increase in language model capacity does not invariably result in performance gains and may even cause overfitting to textual data. Furthermore, \method{} demonstrated notable enhancements over CLOSE across all tested backbones except one, confirming the robustness and superior performance of our proposed method.

\noindent \textbf{The Effects of Dataset Scale.} Text data can be generated and collected with relative ease. To analyze the impact of text data scale on model performance, we conducted experiments summarized in Fig.~\ref{fig:data_scale}. The results demonstrate a consistent trend: larger datasets correspond to improved model performance. This observation underscores the importance of the dataset scale in optimizing model effectiveness, further affirming the value of our \method{} in leveraging extensive text data for improved results.

\noindent \textbf{Vision Driven Language Tasks.}
Initially designed for tasks where language guides visual understanding, our method, \method{}, has also excelled in tasks that use contrastive loss models to bridge different modalities, like the CLAP~\cite{laionclap2023} model, which links audio and text. We tested its versatility by flipping our approach to vision-driven language tasks. In this new setup, models learn from images and are tested using text. We inversed the task settings such as S-Cap, VQA-E, and VE into single text captioning (S-TCap), text-based question answering (TQA), and text entailment (TE), respectively, as shown in Tab.~\ref{table:switch_training}. Both CLOSE and \method{} outperformed the models that did not use noise augmentation. In particular, \method{} improved the CIDEr score by 1 point in S-TCap and increased the accuracy of the TE by 1.2 points, with a minor 0.1-point decrease in the accuracy of the TQA. These findings underscore \method{}'s adaptability and strong performance across different contrastive loss settings for zero-shot cross-modal tasks, showcasing its broad utility.

\section{Conclusion}
In this work, we proposed an effective adaptive noise injection method \method{}, tailored for language-driven visual tasks. \method{} strategically enhances text features by perturbing them into various domains, meticulously preserving content integrity. Besides, our refined noise injection strategy judiciously selects high-quality augmentations, thereby indirectly broadening the scope of injection and augmenting domain diversity.
This dual-faceted approach results in measurable advancements across various visual tasks and network architectures.

The superiority of our \method~over existing methods is evident, although it does encounter challenges with certain intricate visual features. As illustrated in Fig.~\ref{fig:fail_cases}, the method sometimes struggles to disentangle and interpret nuanced details or distinguish between similar foreground and background elements. However, we are confident that the exploration of language-driven visual tasks will catalyze the development of more sophisticated techniques. Moreover, this work is not only a leap forward in bridging text and vision but also holds the potential for application across other modalities, such as point clouds \cite{Afham2022-ll}, videos \cite{Xu2021-sg}, and audio \cite{Elizalde2022-qv, Guzhov2021-lj, Wu2021-dd}. We are confident that the findings from this research will contribute to the development of more effective methods in the field. Additionally, we are interested in exploring semi-supervised approaches, where a limited number of image-text pairs could simplify the task and enhance its practicality for real-world scenarios.
\\
\textbf{\textit{``A data scientist is tasked with developing a low-cost surgical VQA system for a 2-month workshop. She collaborates with a medical expert to generate 5,000 scenario descriptions and QA pairs using ChatGPT, followed by manual refinement. This process takes 2 days and costs under \$2,000. Using the language-driven method \method{}, combined with a surgical vision-language aligned model from Hugging Face, the model achieves adaptability across varied contexts, completing the project in one week.''
}}
{
    \small
    \bibliographystyle{ieeenat_fullname}
    \bibliography{main}
}

\end{document}